\documentclass{ecai}
\usepackage{times}
\usepackage{amsmath}
\usepackage{graphicx}
\usepackage{latexsym}

\begin{document}

\title{On The Plurality of Graphs}

\author{Nicole Fitzgerald*\institute{Microsoft Research, Montreal, nifitzge@microsoft.com} \and Jacopo Tagliabue*\institute{Coveo Labs, New York, jtagliabue@coveo.com} }

\maketitle
\bibliographystyle{ecai}

\begin{abstract}
    We conduct a series of experiments designed to empirically demonstrate the effects of varying the structural features of a multi-agent emergent communication game framework. Specifically, we model the interactions (edges) between individual agents (nodes) as the structure of a graph generated according to a series of known random graph generating algorithms. Confirming the hypothesis proposed in \cite{Fitzgerald2019ToPI}, we show that the two factors of variation induced in this work, namely 1) the graph-generating process and 2) the centrality measure according to which edges are sampled, in fact play a significant role in determining the dynamics of language emergence within the population at hand.
\end{abstract}

\section{INTRODUCTION}
The emergence of communication between agents (humans, animals, robots) is a special case of \textit{agreement by convention}; however, an explicit agreement on word meaning seems to be ruled out by the very nature of language. ``We can hardly suppose", writes Russell, "a parliament of hitherto speechless elders meeting together and agreeing to call a cow a cow and a wolf a wolf''~\cite{Russell1921-RUSTAO-3}~. Following the pioneering work in \cite{lewis1969convention}, the question of how language conventions become established in a population has been broached within different disciplines and with different tools~\cite{skyrms1996evolution,Millikan2005-MILLAB}. 

Following the successes in natural language modelling and natural language generation within the deep learning tradition \cite{10.5555/944919.944966}, the study of language \textit{emergence} between deep neural agents has recently become an active area of research in the intersection of these communities. This research extends the theoretical results presented in \cite{Fitzgerald2019ToPI}, leveraging known random graph-generating algorithms to construct communication networks of neural agents mimicking those found in real-world settings. In particular, we hypothesize that both learning efficiency and communicative success vary if agents are not living in "a vacuum" with unrealistic meeting strategies, but instead are embedded in (and so, constrained by) a detailed network structure. Further, we hypothesize that the choice of network structure bears heavily on these two axes, with certain structural properties facilitating task success whilst others inhibit it. 

Our experiments suggest that the particular connectivity of "scale-free" networks has a significant boosting effect in language emergence, similar to what observed in the viral spread characteristic of social, communication, and contagion networks. \cite{PhysRevLett.86.3200}. More generally, however, these results serve to substantiate the broader hypothesis that \textit{emergent phenomena is governed by structure} --- an idea to be both carefully considered and further explored in the study of multi-agent learning and evolutionary systems.
\section{BACKGROUND}    
\paragraph{Emergent communication.} The study of emergent communication between neural agents is a growing area of interest within the deep learning community. Though a number of variants have been presented over the course of the recent time period, the central focus of this line of study centres around inducing ground-up language acquisition between two agents, which are themselves typically parameterized by neural networks. 

The canonical setting employed in this language acquisition process employs the Lewis Signalling Game schema, first depicted in abstract in \cite{lewis1969convention} and popularized in the deep learning setting in \cite{lazaridou2018emergence}.

In the discrete setup, a sender agent is presented with a target image $x_t$ while its listener agent is presented with a set of candidate images $X = \{x_1, ..., x_n\}$, $x_t \in X$ which contains the target image and $|X| - 1$ distractor images. During gameplay, the sender selects symbols from a vocabulary $W$ to construct a message $m = \{w_1, ..., w_m\}$, $w_i \in W$ which serves as a description of the target image. Given $m$, the receiver must then identify the target from its set of candidates. \textit{Communicative success} is defined as the correct identification of the target image by the listening agent. The game may alternately be played in the continuous setting, in which the messages generated by the sender agent take the form of fixed-size continuous vectors, but we opt here to lend our focus solely to the discrete case.

\begin{figure}[htp]
    \centering
    \includegraphics[width=6.00cm]{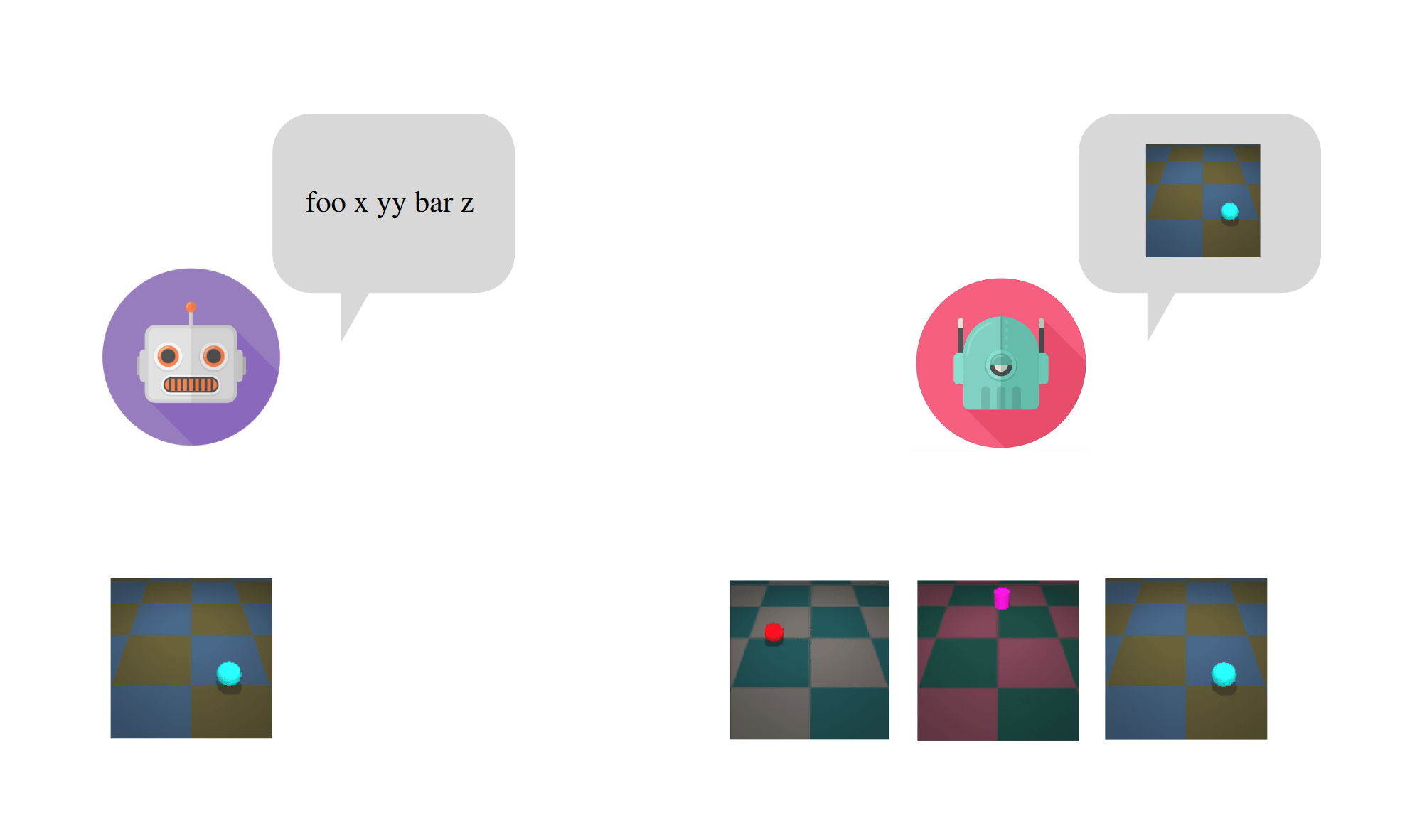}
    \caption{Lewis signalling game between two agents. The sender agent (left) is presented with a target image and conveys a message to the receiver agent (right), which must then attempt to distinguish the target image from its set of candidate image. }
    \label{fig:lewis_signalling_game}
\end{figure}

\section{RELATED WORKS}

\paragraph{Population-based learning.} While the majority of work in this area has opted to focus on the emergence of language use between pairs of agents \cite{lazaridou2018emergence}, a small number of studies exploring language emergence in the context of a \emph{population} have begun to surface. 

Notably, \cite{Fitzgerald2019ToPI} examines the extent to which representations contained within the image-encoding component of the agent architecture are affected by the size of the population within which a given agent is trained, suggesting that the population-based setting acts as an implicit regularizer against pair specific idiosyncrasies that tend to emerge in communication games. 

In a similar vein, \cite{tielemanshaping} demonstrates increasing reduction of idiosyncratic latent codes between the encoder and decoder components of a conventional autoencoder architecture proportional to the size of the populations of encoders and decoders from which these components are randomly sampled.

Finally, \cite{graesser2019emergent} offers an extensive overview concerning the effects of connectivity manipulation on both intra- and inter- population learning in communication games, observing that the degree of connectivity both \textit{within} a population and \textit{to} neighbouring populations strongly affects language convergence between agents and populations respectively.

We distinguish this particular work from prior works in that, to our knowledge, we offer the first explicit study isolating the effects of network topology on the emergence of language between neural agents. Additionally, this work is the first to explore the idea of modelling the population dynamics of these agents with graph algorithms motivated by real-world network structures. In all previous works, agents were paired together either according to an arbitrary random process or in a fully-connected manner. While the former case has been found to occasionally handicap the learning capacity of agents within the population, the latter suffers from quadratic scaling in training time proportional to population size, making it intractable for anything beyond extremely small populations.  

\paragraph{Network-based simulations.} Real-world agents can be heterogeneous, learn over time and be situated in specific environments that constrain their actions; as it is often hard for closed-form solutions to deal with these features, explicit simulations of complex systems are a powerful tool to gain insights on population dynamics that are too complex to be modelled analytically (e.g. the seminal \cite{doi:10.1080/0022250X.1971.9989794}). Agent-based models \cite{Page+Miller} can incorporate spatial constraints, difference in agents abilities and preferences and the evolution of specific traits in the population of interests (\cite{ab_model_2016}, \cite{10.2307/j.ctvcm4gjh}). When coupled with network theory \cite{10.5555/1809753}, simulations allow to investigate how properties of the underlying networks affect the target phenomenon (e.g. \cite{ANDERSON201819} in biology, \cite{Suzuki2015} in economics, \cite{Primiero2017} in trust and information spreading). To the best of our knowledge, this work is the first explicit network-based simulation of a population of neural agents collectively trying to learn a language.

\section{EXPERIMENTAL SET-UP}
\subsection{Population Topology}

\begin{figure}[htp]
    \centering
    \includegraphics[width=2.80cm]{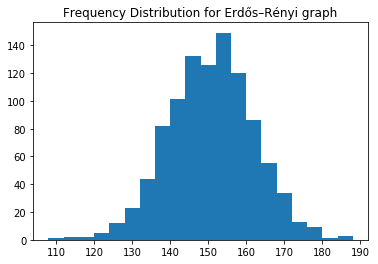}
    \includegraphics[width=2.80cm]{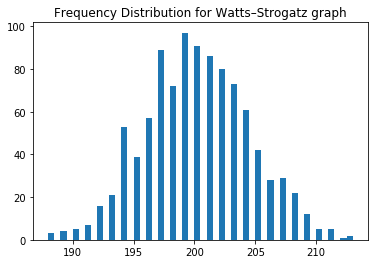}
    \includegraphics[width=2.80cm]{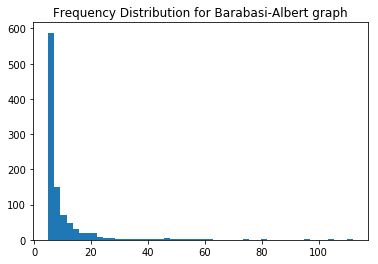}
    \caption{Degree distribution for the three graph topologies under consideration ($n=1000$). From left to right: \textbf{ER}, \textbf{WS} and \textbf{BA}. Graphs are generated with the \textit{networkx} library \cite{networkx}.}
    \label{fig:degree_distributions}
\end{figure}

\paragraph{Graph generation.} Extending the work of \cite{Fitzgerald2019ToPI}, agent interactions do not exist "in a vacuum" --- that is to say, randomly paired, but instead they are embedded in graphs that constraint their interaction. Graphs are usually classified according to their topology \cite{10.5555/1809753}, since two graphs with the same nodes and same number of edges may exhibit drastically different behavior (e.g. how they constrain the spreading of a virus \cite{10.1371/journal.pone.0012948}) depending on their structure. Three different settings are tested choosing three representative topologies, which substantially differ for their degree distribution (Figure \ref{fig:degree_distributions}). Given a graph with $n$ nodes:

\begin{itemize}
    \item \textbf{Erdős–Rényi} graph \cite{erdos59a} (henceforth ER).This type of graph exhibit short average path length (i.e. the distance between any two given nodes is, on average, small), but also low clustering; a graph is generated by connecting the $n$ nodes randomly: in particular, given two nodes, they are connected with probability $p$ independent from every other edge. We consider the ER graph topology as our baseline, in which agent interactions are a function of randomness. 
    \item \textbf{Watts–Strogatz} graph \cite{wattsNature} (henceforth WS). This type of graphs exhibit the ``small world'' property, as it allows for the formation of clusters while retaining short average path length typical of the \textit{ER} model; a graph is generated by first placing the $n$ nodes in a ring and join each node with its $k$ nearest neighbors; existing edges are then replaced and rewired with probability $p$ to obtain the final layout.
    \item \textbf{Barabási–Albert} graph \cite{Albert2001StatisticalMO} (henceforth BA). This type of graph exhibit the power-law distribution often observed in social networks \cite{Barabsi1999EmergenceOS}, in which a few "hubs" will have a disproportionally high number of connections. A graph is generated by adding new nodes, from $1$ to $n$, with a mechanism of preferential attachment, where the probability of being added and connected to $m$ existing nodes $x_1, x_2,...~x_m$ is proportional to the degree of $x_1, x_2,...~x_m$.
\end{itemize}

\paragraph{Sampling methods.}Graphs constrain the interactions of our agents in two important ways: i) we can only sample pairs of agents which are connected in the underlying network, ii) we can make nodes more salient during sampling by modelling the importance of each node in the graph as a function of its \textit{centrality} - the more important the node, the more it will be selected for the game (Section \ref{p:edge_sampling}). 

Centrality measures are a convenient way to identify the most important nodes within a graph (different measures define "importance" in slightly different ways): for our experiments we pick three among the most common measures in network analysis \cite{10.5555/1809753}: \textit{degree} centrality (in which centrality is simply a function of the node's degree), \textit{betweeness} centrality \cite{Brandes01afaster} (in which centrality measures the number of shortest paths that pass through a node), \textit{PageRank} centrality \cite{Brin98theanatomy} (in which centrality depends on the quantity and quality of the edges of a node).

\subsection{Training}
For each experiment, we instantiate a population of \textit{n} agents according to a graph generation algorithm \textit{G} and collect an ordered set of agent pairs, or edges, according to a sampling procedure \textit{S}. Following the training procedure delineated in \cite{Fitzgerald2019ToPI}, we iterate over the set of agent pairs, training each pair for a fixed number of steps over mini-batches of communication games before returning each of them to the population. At each time step, we randomly assign one agent to play the role of the sender and the other to take on the role of the receiver, passing the agent's encoder output through the appropriate head. As such, each agent trains for roughly half of its total steps as a sender agent and half as a receiver agent. Below, we detail the specifics of the graph generation, sampling, pair-based training processes in turn. 

\paragraph{Graph Generation.}\label{p:graph_generation}
To control for the total number of nodes and edges among all the graphs in the tested topologies, we sought to ensure graphs have a constant number of nodes and edges, as standard input parameters to generate graphs typically do not ensured guaranteed cardinality of the edge set. We use the following three formulas to calculate the value of free parameters to obtain a graph with $n$ nodes and $e$ edges (please note that i) for \textit{ER} we have that the \textit{expected} number of edges is $e$, ii) for \textit{BA}, we pick the smaller integer).

\begin{equation}
    \tag{ER}
    p = \frac{e}{n^2 / 2}
\end{equation}

\begin{equation}
    \tag{WS}
    k = \frac{e * 2}{n}
\end{equation}

\begin{equation}
    \tag{BA}
    f(m) = m^2 - nm + e
\end{equation}

\paragraph{Edge sampling.}
\label{p:edge_sampling}
Agents are sampled from the population in a measure proportional to their centrality; when the underlying graph is generated, centrality is calculated for each node; scores (which depend on the centrality measure adopted) are passed through a softmax function and finally nodes are sampled according to the resulting score.

\paragraph{Pair-based training.}
We set each agent pair to train together for a total of 1 024 00 games, which is decomposed into a set of mini-batches, each of size 32. The full optimization details can be found in \ref{subsec:learning}. As noted in preceding paragraphs, each agent spends roughly half of its time in the sender setting and half in the receiver setting. 

Upon the conclusion of the training cycle of a given pair of agents, the agents are then returned to the population in their trained state, remaining as is until the next occasion they are selected. This set-up implicitly mimics the continual, or life-long, learning \cite{chen2018lifelong} set-up, in that from the point of view of a single agent this method decomposes into a single learning problem in which the agent faces a stream of input data originating from a continuously changing distribution. 

\subsection{Evaluation} 
In the evaluation phase, we generate 10 randomly selected pairs of agents from within a trained population, regardless of their true relationship within the underlying graph structure. In a similar fashion to the training procedure, we train each pair in this new collection for a total of 320 000 steps. In the case that a particular agent is sampled more than once, we reset its parameters to those obtained in the training phase, ensuring that no information from the evaluation phase is carried over.

The games generated in the evaluation phase sample images from a held-out set of images originating from the same distribution as the training set (see \ref{subsec:data}) 

Our evaluation set up is designed to monitor and analyze the ability of a trained agent to quickly adapt to an effective communication protocol with a new partner, regardless of whether or not the two of them have communicated in the past. As such, we take both the sample efficiency and the overall performance of these learning curves into consideration in the analysis of our evaluation results. 

\subsection{Data} \label{subsec:data}
\paragraph{Training} As in \cite{lazaridou2018emergence} and \cite{Fitzgerald2019ToPI}, we construct our set of training games from a set of 4000 synthetic images of geometric shapes, generated by the Mujoco physics engine. A single game is generated by randomly sampling a set of $X$ images from the dataset and subsequently sampling a single image $x_t \in X$ as the target image. 

\paragraph{Evaluation} The test set is similarly constructed over a collection of 1000 images from the same source.

\begin{figure}[htp]
    \centering
    \includegraphics[width=2cm]{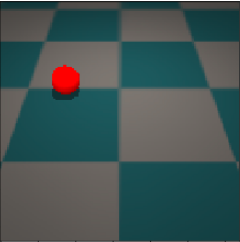}
    \includegraphics[width=2cm]{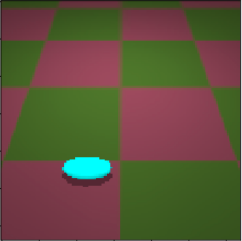}
    \includegraphics[width=2cm]{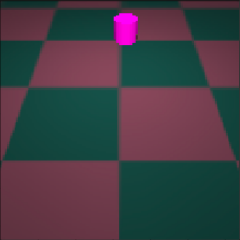}
    \caption{Training examples generated by the Mujoco physics engine. There exist three main factors of variation, namely the \textit{color} and \textit{shape} of the item pictured, in addition to the color scheme of the underlying tiling.}
    \label{fig:results}
\end{figure}

\section{IMPLEMENTATION DETAILS}
\subsection{Agents}
In this work, we adopt a joint speaker-listener architecture in which each agent is comprised of a single CNN encoder and two policy heads, one of which is used as the \emph{sender} head and the other as the \emph{receiver} head. That is to say, we equip each agent to act as either a speaker or a listener given the context at hand, and optimize the image encoder with respect to both of the respective policies. 

\begin{figure}[htp]
    \centering
    \includegraphics[width=9.00cm]{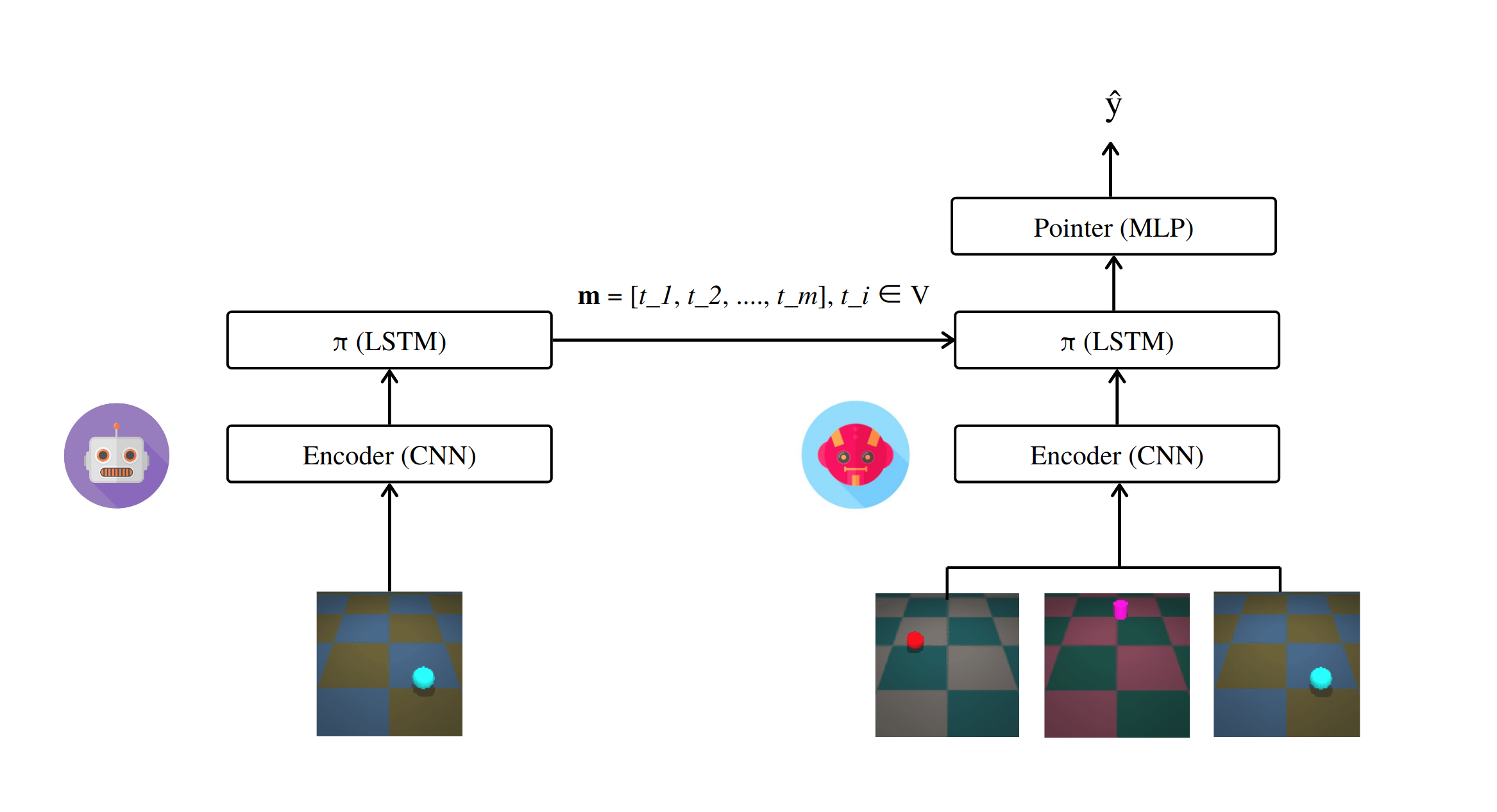}
    \caption{The agent architecture employing the sender head (left) and the agent architecture employing the receiver head (right). For a single agent, there exists a central CNN encoder. The output of this encoder flows through to either the sender head or the receiver head, according to the agent's role at a given time step.}
    \label{fig:lewis_architecture}
\end{figure}

The CNN image encoder $f(\theta_{f}, x_i)$ is composed of four convolutional layers, each of which is followed by a batch-normalization operation \cite{} and a ReLU activation. The output of the final convolutional layer is passed through a single linear layer to produce a fixed-size dense representation vector $u_i$ corresponding to an image $x_i$. In the sender setting, the agent encodes a single image $x_t$, producing representation vector $u_t$. In the receiver setting, the agent encodes each image in the candidate set individually, producing a set of representation vectors $U = \{u_{i} = f(\theta^{L}, x_i) | x_i \in X\}$.

The sender head is a recurrent policy generated by a single-layer LSTM $h^{S}(\theta_{h}^{S}, u_t)$ \cite{hochreiter1997long} decoder which takes the representation vector produced by the encoder as its input and produces a variable length message via sampling discrete tokens from a vocabulary V according to the LSTM hidden state at \emph{t}.

The receiver head encodes message \textbf{m} via a single-layer LSTM $h^L(\theta_{h}^L, z)$, where $z$ is the embedded vector representation of \textbf{m}.  The listener then selects an image $t'$ by sampling from a Gibbs distribution generated via the dot product of $z$ and all encoded images $u \in U$. 

\subsection{Learning}\label{subsec:learning}
The shared objective function (Eq. \ref{eq:3}) optimized by a given speaker-listener pair may be decomposed as the sum of the individual speaker (Eq.\ref{eq:1}) and receiver (Eq.\ref{eq:2}) losses:

\begin{equation} \label{eq:1}
    \mathcal{L}(\theta_{f}^S, \theta_{h}^S) = (\sum_{l=1}^L \log p_{{\pi}^{S}}(m^{l}_{t}|m^{<l}_t, u_{t})) 
\end{equation}

\begin{equation} \label{eq:2}
    \mathcal{L}(\theta_{f}^L, \theta_{h}^L) =  \log p_{{\pi}^{L}}(t'|z, U))
\end{equation}

\begin{equation} \label{eq:3}
    \mathcal{L}(\theta) = ((R(t')-b) \cdot (L(\theta_{f}^S, \theta_{h}^S) + L(\theta_{f}^L, \theta_{h}^L)- H_S)   
\end{equation}

where $b$ is a baseline variance reduction term that we simply set to $b = \frac{1}{N} \sum r(\tau)$, $R(t')$ is the reward, which is $1$ if the predicted target is correct (i.e. $t=t'$) and 0 otherwise, and $z$ is the encoding of the message \textbf{m} computed by the listener LSTM. The entropy term $H_S$ corresponding to the entropy of the speaker's policy is a regularization term added to encourage exploration \cite{mnih2016asynchronous}.

Given the discrete nature of the messages, we estimate model parameters via the REINFORCE update rule \cite{williams1992simple}. The set of parameters optimized at a single time step is a function of the role played by the agent at that particular $t$; specifically, in the sender setting, we optimize only the parameters pertaining to the LSTM decoder $h^{S}(\theta_{h}^{S}, u_t)$ and freeze those pertaining to the receiving LSTM encoder $h^L(\theta_{h}^L, z)$. The converse is true in the receiver setting. In both cases, we update the parameters contained within the central image encoder $f(\theta_{f}, x_i)$. As previously noted, all agent parameters are randomly initialized at the beginning of the evaluation phase.

\subsection{Hyperparameters}
In this work, we employ the following set of hyperparameters: we set the size of each population to be $n = 16$ agents, we configure our parameter control mechanisms (see \ref{p:graph_generation}) to generate 32 edges within the graph and similarly employ our sampling algorithm of choice to select 32 edges according to the given centrality measure. The size of the vocabulary, i.e. $|V| = 20$ and the message length $L$ is at most 5. We set the dimension of the embedding matrix used to embed the message tokens to be size 32. The dimensionality of the speaker and listener LSTM hidden states is 64. 

We multiply the speaker entropy term by a coefficient $\alpha = 0.1-|(R(t') - b| \cdot 0.1$ while $speaker\_steps < 1 000 000$ and $0.01$ otherwise.

We train and evaluate the agents with a set of 4 candidate images, i.e. $|X| = 4$.

\section{RESULTS}
\begin{figure*}[htp]
    \centering
    \includegraphics[width=13.00cm]{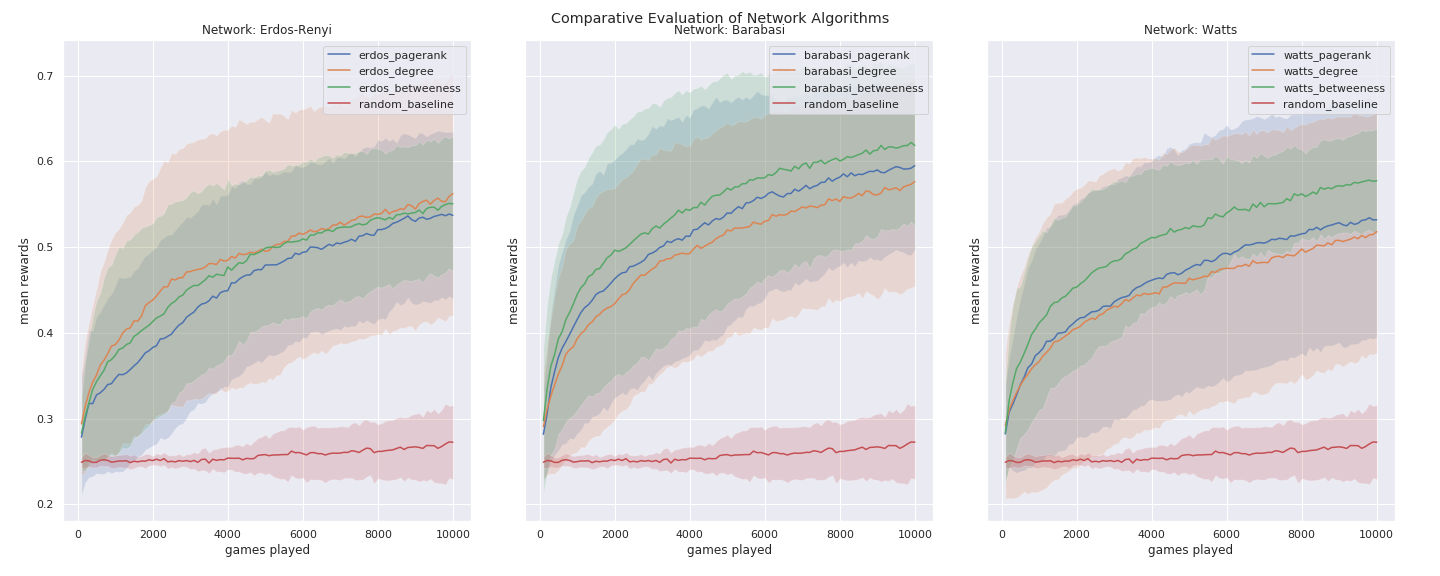}
    \caption{Mean rewards on 10000 games for different topologies vs random pairing baseline. From left to right: \textbf{ER}, \textbf{WS} and \textbf{BA}. Trend line is the \textit{mean} reward, lightly colored bands represent one standard deviation.}
    \label{fig:graph_mean_reward}
\end{figure*}

\begin{figure*}[htp]
    \centering
    \includegraphics[width=13.00cm]{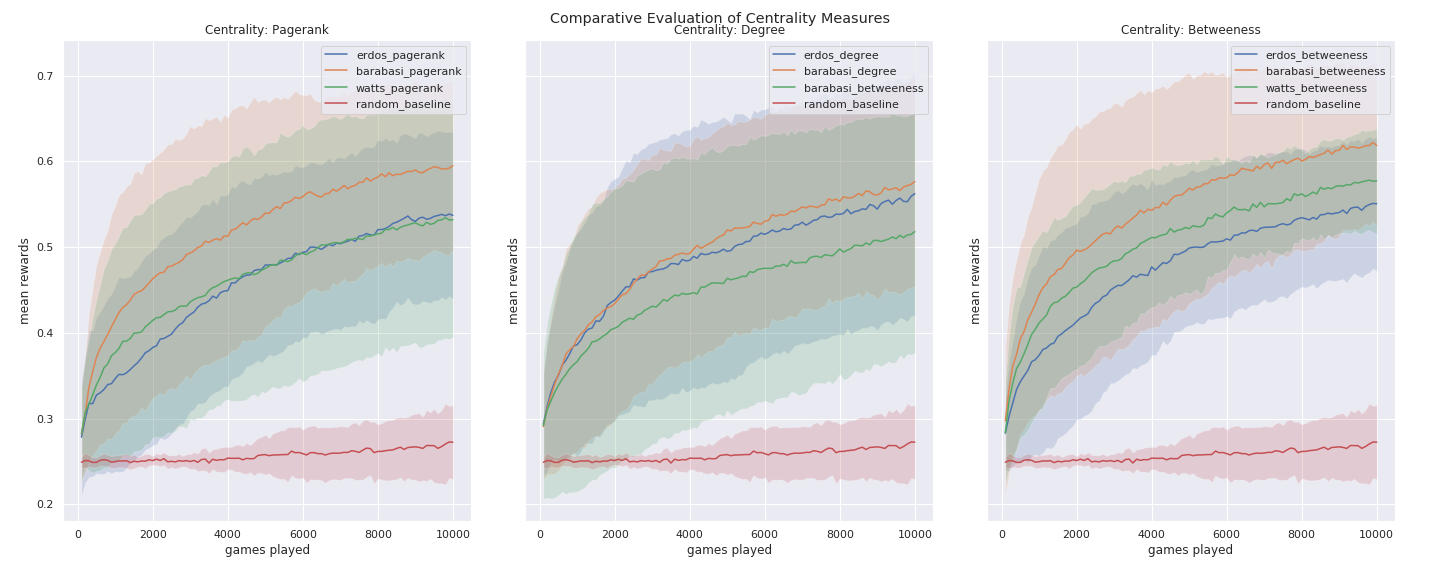}
    \caption{Mean rewards on 10000 games for different centrality measures vs random pairing baseline. From left to right: \textbf{PageRank}, \textbf{Degree} and \textbf{Betweeness}. Trend line is the \textit{mean} reward, lightly colored bands represent one standard deviation.}
    \label{fig:centrality_mean_reward}
\end{figure*}

 Figure \ref{fig:graph_mean_reward} shows the mean reward for the agents across graph topologies. The first result is that the new graph setting consistently outperforms the baseline score obtained through random pairing of agents; in particular, among all graphs, \textit{BA} - featuring both clustering through "hubs" and small average path length - has the best performances: agents playing the game on \textit{BA} learn more (peak reward is higher) and faster (reward shows a steeper increase in the first 1000 games). As shown in Figure \ref{fig:centrality_mean_reward} - which displays graph performances across different centrality measures - \textit{BA} is consistently the best topology, with the biggest difference in the \textit{betweeness} case.

The increase in performances produced by \textit{BA} is consistent with previous findings in the literature on network phenomena (\cite{NEKOVEE2007457}, \cite{10.1145/956750.956769}, \cite{PhysRevLett.86.3200}). To explain the mean reward curves we observed, two hypotheses are put to test: i) central agents in the network ("hubs") help spreading the knowledge and act as a catalyst throughout the population by quickly bringing everybody up to speed (so to speak); ii) central agents exploit their position during sampling/training to accumulate a disproportional amount of knowledge and outperform everybody else - the mean reward we observe would then be due to the fact that few agents perform exceptionally well, and the vast majority of agents are mediocre. 

\begin{figure*}[htp]\label{fig:agent_results}
    \centering
    \includegraphics[width=16.00cm]{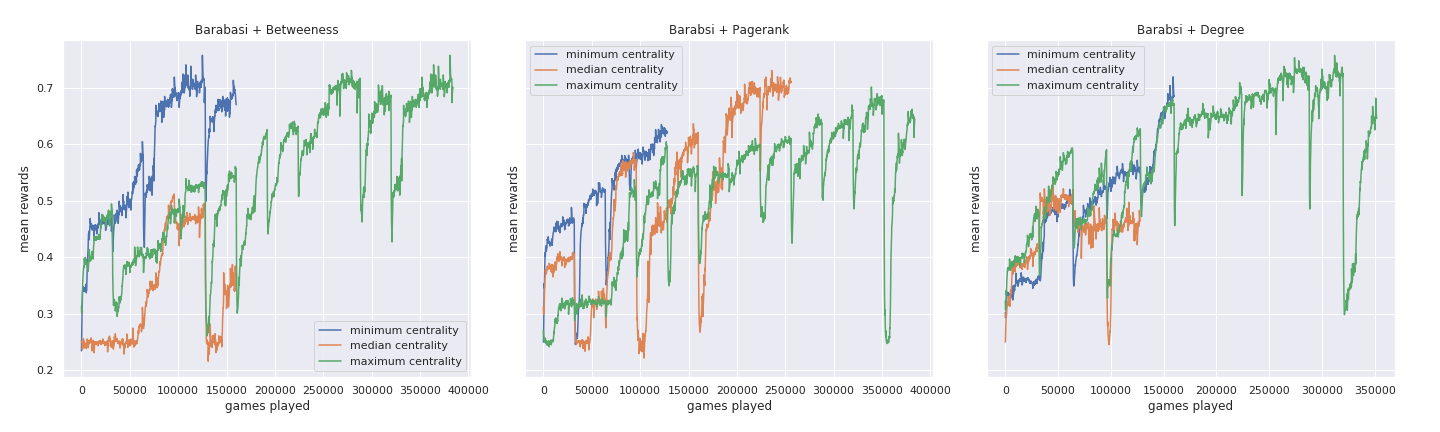}
    \caption{Training trajectory (mean reward) on \textit{BA} across centrality measures for three agents with different centrality rank: \textit{minimum} (blue) vs \textit{median} (orange) vs \textit{maximum} (green). Please note that the trajectory for each agent is cut at the last point available in \textit{all} the (random seeded) runs, to avoid averaging over incomparable sets.}
    \label{fig:centrality_train_curves}
\end{figure*}

To distinguish between (i) and (ii), we report reward trajectories for agents training on \textit{BA} in Figure \ref{fig:centrality_train_curves}, by plotting an agent with minimal centrality, an agent with median centrality and an agent with maximum centrality. 

The trajectories of the three agents show a similar progressions and best score, favoring the first of the two hypotheses: the scale-free graph makes possible for \textit{all} agents in the network to accelerate learning. The surprising sample efficiency of the minimally central agent (\textit{blue} line) may be explained by the small variety of its neighborhood: we look forward to investigating the interplay of centrality and generalization abilities in subsequent work.

All experiments are averaged over three random seeds. 

\section{CONCLUSIONS AND FUTURE WORK}
This work presents the first network-based model for neural agents playing a signaling game: by playing the game on a graph, agents' interactions and sampling are determined by the underlying topology; we showed that learning efficiency and mean performance are significantly affected by the graph and demonstrated that all the tested graph-based models significantly outperform a baseline where agents play the game "in a vacuum", i.e. are paired and ordered randomly, confirming the prediction put forward in \cite{Fitzgerald2019ToPI}. 

This line of work, namely that concerned with exploring the emergence of language within multi-agent populations, is an extremely diverse and dynamic field of study, encompassing a number of related fields - machine learning, linguistics, graph theory, information theory, social dynamics,  and evolutionary mechanisms. As such there exist a number of avenues to further this body of research, which we present here. 

Our results, though preliminary, accentuate a core idea in multi-agent interactions --- namely, \textit{structure matters}. The manner in which agent interactions are organized in turn shapes the nature of the interactions that arise between said agents. Within our context, the word \textit{interactions} encompasses a number of phenomena --- in particular, we point to learning dynamics and rules of convention as phenomena of note. To this end, we are interested in further exploring the extent to which structure affects emergent and systemic phenomena in the multi-agent setting. There are two axes of interest at play: first, we ask, can we identify optimal or desirable structures for multi-agent learning? As an appendage to this question, we ask, can we formalize the properties that make said structures optimal or desirable? Second, can we \textit{learn} these structures --- that is, can we eliminate the need for hand-engineering population structures and allow structures to emerge as a function of agent choice? Do these choices converge towards optimal solutions?

Aligning with the interests of both the linguistics community and much of the emergent communication community, it might prove a worthwhile exercise to more closely examine the linguistic structure of the emerged communication protocols. Inspired by evidence of groups of universal linguistic order properties \cite{hahn2020universals} emerging in natural languages, it would be interesting to examine the extent to which (if at all) the structure of the underlying network determines network-specific linguistic properties of the emerged language across all seeds.

Following previous, non-neural work, on the emergence of linguistic conventions through iterated games \cite{skyrms1996evolution}, a natural extension to the current setting would be to induce some degree of evolutionary dynamics in the population of learning agents.

Following the intuitions posited by \cite{lazaridou2018emergence} and \cite{Fitzgerald2019ToPI}, a simple and natural extension of this work might involve further analysis of the learned representations present within the agents' image-encoders, investigating the degree to which they manage to encode salient features and their use for downstream tasks. 

\ack We would like to thank the referees for their comments, which
helped improve this paper considerably

\bibliography{ecai}

\begin{thebibliography}{10}

\bibitem{Albert2001StatisticalMO}
R{\'e}ka Albert and Albert-L{\'a}szl{\'o} Barab{\'a}si, `Statistical mechanics
  of complex networks', {\em ArXiv}, {\bf cond-mat/0106096}, (2001).

\bibitem{ANDERSON201819}
Taylor~M. Anderson and Suzana Dragićević, `Network-agent based model for
  simulating the dynamic spatial network structure of complex ecological
  systems', {\em Ecological Modelling}, {\bf 389},  19 -- 32, (2018).

\bibitem{Barabsi1999EmergenceOS}
Barab{\'a}si and Albert, `Emergence of scaling in random networks', {\em
  Science}, {\bf 286 5439},  509--12, (1999).

\bibitem{10.5555/944919.944966}
Yoshua Bengio, R\'{e}jean Ducharme, Pascal Vincent, and Christian Janvin, `A
  neural probabilistic language model', {\em J. Mach. Learn. Res.}, {\bf
  3}(null),  1137–1155, (March 2003).

\bibitem{Brandes01afaster}
Ulrik Brandes, `A faster algorithm for betweenness centrality', {\em Journal of
  Mathematical Sociology}, {\bf 25},  163--177, (2001).

\bibitem{Brin98theanatomy}
Sergey Brin and Lawrence Page, `The anatomy of a large-scale hypertextual web
  search engine', in {\em COMPUTER NETWORKS AND ISDN SYSTEMS}, pp. 107--117,
  (1998).

\bibitem{chen2018lifelong}
Zhiyuan Chen and Bing Liu, `Lifelong machine learning', {\em Synthesis Lectures
  on Artificial Intelligence and Machine Learning}, {\bf 12}(3),  1--207,
  (2018).

\bibitem{10.1371/journal.pone.0012948}
Nicholas~A. Christakis and James~H. Fowler, `Social network sensors for early
  detection of contagious outbreaks', {\em PLOS ONE}, {\bf 5}(9),  1--8, (09
  2010).

\bibitem{erdos59a}
P.~Erd\"{o}s and A.~R\'{e}nyi, `On random graphs i', {\em Publicationes
  Mathematicae Debrecen}, {\bf 6},  290, (1959).

\bibitem{Fitzgerald2019ToPI}
Nicole J~M Fitzgerald, `To populate is to regulate', {\em ArXiv}, {\bf
  abs/1911.04362}, (2019).

\bibitem{10.2307/j.ctvcm4gjh}
Herbert Gintis, {\em Game Theory Evolving: A Problem-Centered Introduction to
  Modeling Strategic Interaction - Second Edition}, Princeton University Press,
  rev - revised, 2 edn., 2009.

\bibitem{graesser2019emergent}
Laura Graesser, Kyunghyun Cho, and Douwe Kiela, `Emergent linguistic phenomena
  in multi-agent communication games', {\em arXiv preprint arXiv:1901.08706},
  (2019).

\bibitem{networkx}
Aric Hagberg, Pieter Swart, and Daniel Chult, `Exploring network structure,
  dynamics, and function using networkx', (01 2008).

\bibitem{hahn2020universals}
Michael Hahn, Dan Jurafsky, and Richard Futrell, `Universals of word order
  reflect optimization of grammars for efficient communication', {\em
  Proceedings of the National Academy of Sciences}, {\bf 117}(5),  2347--2353,
  (2020).

\bibitem{hochreiter1997long}
Sepp Hochreiter and J{\"u}rgen Schmidhuber, `Long short-term memory', {\em
  Neural computation}, {\bf 9}(8),  1735--1780, (1997).

\bibitem{10.1145/956750.956769}
David Kempe, Jon Kleinberg, and \'{E}va Tardos, `Maximizing the spread of
  influence through a social network', in {\em Proceedings of the Ninth ACM
  SIGKDD International Conference on Knowledge Discovery and Data Mining}, KDD
  ’03, p. 137–146, New York, NY, USA, (2003). Association for Computing
  Machinery.

\bibitem{lazaridou2018emergence}
Angeliki Lazaridou, Karl~Moritz Hermann, Karl Tuyls, and Stephen Clark,
  `Emergence of linguistic communication from referential games with symbolic
  and pixel input', {\em arXiv preprint arXiv:1804.03984}, (2018).

\bibitem{lewis1969convention}
David Lewis, `Convention', {\em Mass.: Harvard UP}, (1969).

\bibitem{Millikan2005-MILLAB}
Ruth~Garrett Millikan, {\em Language: A Biological Model}, Oxford: Clarendon
  Press, 2005.

\bibitem{mnih2016asynchronous}
Volodymyr Mnih, Adria~Puigdomenech Badia, Mehdi Mirza, Alex Graves, Timothy
  Lillicrap, Tim Harley, David Silver, and Koray Kavukcuoglu, `Asynchronous
  methods for deep reinforcement learning', in {\em International conference on
  machine learning}, pp. 1928--1937, (2016).

\bibitem{ab_model_2016}
Akira Namatame and Shu-Heng Chen, {\em Agent-based modelling and network
  dynamics}, Oxford University Press, 2016.

\bibitem{NEKOVEE2007457}
M.~Nekovee, Y.~Moreno, G.~Bianconi, and M.~Marsili, `Theory of rumour spreading
  in complex social networks', {\em Physica A: Statistical Mechanics and its
  Applications}, {\bf 374}(1),  457 -- 470, (2007).

\bibitem{10.5555/1809753}
Mark Newman, {\em Networks: An Introduction}, Oxford University Press, Inc.,
  USA, 2010.

\bibitem{Page+Miller}
Scott~E Page and John~H. Miller, {\em Complex Adaptive Systems: An Introduction
  to Computational Models of Social Life}, Princeton University Press, 2007.

\bibitem{PhysRevLett.86.3200}
Romualdo Pastor-Satorras and Alessandro Vespignani, `Epidemic spreading in
  scale-free networks', {\em Phys. Rev. Lett.}, {\bf 86},  3200--3203, (Apr
  2001).

\bibitem{Primiero2017}
Giuseppe Primiero, Franco Raimondi, Michele Bottone, and Jacopo Tagliabue,
  `Trust and distrust in contradictory information transmission', {\em Appl
  Netw Sci}, {\bf 2}, (2017).

\bibitem{Russell1921-RUSTAO-3}
Bertrand Russell, {\em The Analysis of Mind}, Duke University Press, 1921.

\bibitem{doi:10.1080/0022250X.1971.9989794}
Thomas~C. Schelling, `Dynamic models of segregation', {\em The Journal of
  Mathematical Sociology}, {\bf 1}(2),  143--186, (1971).

\bibitem{skyrms1996evolution}
B.~Skyrms and Cambridge~University Press, {\em Evolution of the Social
  Contract}, Cambridge University Press, 1996.

\bibitem{Suzuki2015}
Aruka~Y. Suzuki~Y., Namatame~A., `Agent-based modeling of economic volatility
  and risk propagation on evolving networks', in {\em Proceedings of the 18th
  Asia Pacific Symposium on Intelligent and Evolutionary Systems}, (2015).

\bibitem{tielemanshaping}
Olivier Tieleman, Angeliki Lazaridou, Shibl Mourad, Charles Blundell, and Doina
  Precup, `Shaping representations through communication', {\em OpenReview},
  (2018).

\bibitem{wattsNature}
Duncan~J. {Watts} and Steven~H. {Strogatz}, `Collective dynamics of small-world
  networks', {\em Nature}, {\bf 393}(6684),  440--442, (June 1998).

\bibitem{williams1992simple}
Ronald~J Williams, `Simple statistical gradient-following algorithms for
  connectionist reinforcement learning', {\em Machine learning}, {\bf 8}(3-4),
  229--256, (1992).

\end{thebibliography}
\end{document}